\documentclass{article}

\makeatletter
\newcommand\deletedfootnote[1]{%
    \ifthenelse{\boolean{Changes@optiondraft}}
    {\footnote{\deleted{#1}}}
    {}}
\makeatother

\usepackage[utf8]{inputenc} 
\usepackage[T1]{fontenc}    
\usepackage{hyperref}       
\usepackage{url}            
\usepackage{booktabs}       
\usepackage{amsfonts}       
\usepackage{nicefrac}       
\usepackage{microtype}      
\usepackage{graphicx}

\title{Neural Architecture Search for Inversion}

\author{%
Cheng Zhan, Licheng Zhang, Xin Zhao, Chang-Chun Lee, Shujiao Huang
}
\date{}

\begin{document}

\maketitle

\begin{abstract}
    Over the year, people have been using deep learning to tackle inversion problems, and we see the
framework has been applied to build relationship between recording wavefield and velocity (Yang et al.,
2016). Here we will extend the work from 2 perspectives, one is deriving a more appropriate loss function,
as we know, pixel-2-pixel comparison might not be the best choice to characterize image structure, and
we will elaborate on how to construct cost function to capture high level feature to enhance the model
performance. Another dimension is searching for the more appropriate neural architecture, which can be
viewed as a subproblem within hyperparameter optimization, which is a subset of an even bigger picture,
the automatic machine learning, or AutoML. There are several famous networks, U-net, ResNet (He et al.
2016) and DenseNet (Huang et al., 2017), and they achieve phenomenal results for certain problems, yet
it’s hard to argue they are the best for inversion problems without thoroughly searching within certain
space. Here we will be showing our architecture search results for inversion.
\end{abstract}

\section{Introduction}

Most of seismic data contains confidential information and therefore are not available to the public. There is very little seismic data available to run experiment. In this paper, we will mainly introduce two approaches to generate synthetic data, one is physics driven that utilizes partial differential wave equation to simulate the the process of wave propagation, and another is more machine learning flavor with GAN as the underlying method to create subsurface velocity models.

\section{Methods and Results}

\begin{figure}
  \centering
    \includegraphics[width=0.8\linewidth]{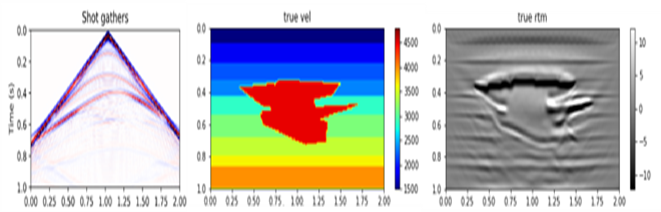}
    \caption{An example of generated shot gathers, velocity, and RTM image.}
    \label{fig:shot_vel_img}
\end{figure}

\begin{figure}
  \centering
    \includegraphics[width=0.8\linewidth]{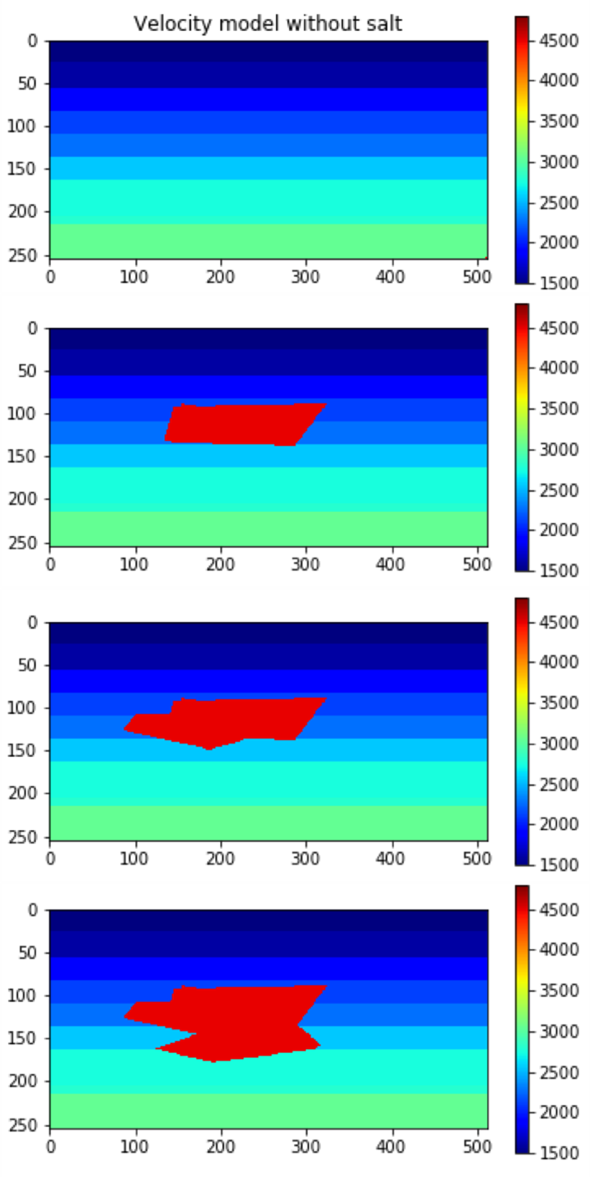}
    \caption{Velocity generation procedure.}
    \label{fig:vp}
\end{figure}

We adopted an open source code generation framework, Devito, (see [1] for more details) to generate the synthetic data including both the wave-fields and reversed time migration (RTM) images. On a higher level. Devito is a symbolic finite difference computational solver and designed to create wave propagation kernels for the application in seismic forward modelling, inversion, and computation fluid dynamic. First, we generated a variety of background layer velocity models followed by inserting complicated salt bodies on top of them. In order to ensure the diversity and richness of the training data, many degrees of randomness were introduced, both in the sediment and salt velocity components. For example, the salt body was a union of several convex hulls, and the number of hulls was a random number; as well as the coordinates of points to create those convex hulls (see Figure \ref{fig:vp} for the step-by-step velocity generation process). Then forward modelling was employed to generate corresponding shot gathers, or recorded seismic wave-field. In the last step, RTM images were produced. Figure \ref{fig:shot_vel_img} shows an example of generated shot gather, underlying velocity, and final product RTM image. 

Besides leveraging the finite difference solver, we also utilized more machine learning flavored, in particular, deep learning framework, GAN, to potentially generate more data. The core idea of GAN is based on the "indirect" training through the discriminator, which itself is also iteratively updated. This basically means that the generator is not trained to minimize the distance to a specific image, but rather to fool the discriminator. This procedure enables the model to learn in an unsupervised fashion. In other words, the generative network generates candidates while the discriminative network evaluates them. The contest operates in terms of statistical distributions. Typically, the generative network learns to map from a latent space to a data distribution, while the discriminative network distinguishes candidates produced by the generator from the true data distribution. The generative network's training objective is to increase the error rate of the discriminative network (see [2] for more details).

\begin{figure}
  \centering
    \includegraphics[width=0.8\linewidth]{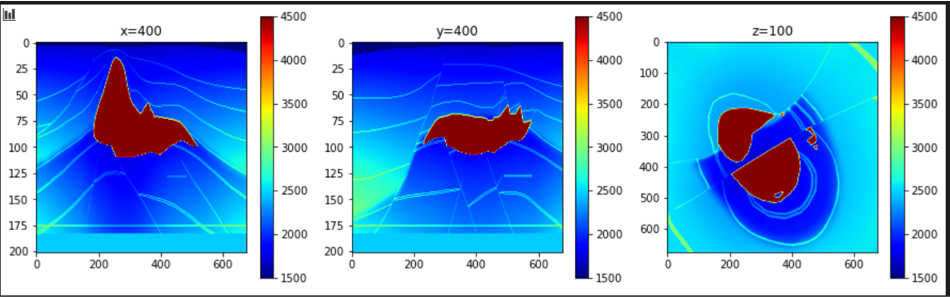}
    \caption{Inline, crossline and top views of the velocity.}
    \label{fig:xyz}
\end{figure}

\begin{figure}
  \centering
    \includegraphics[width=0.8\linewidth]{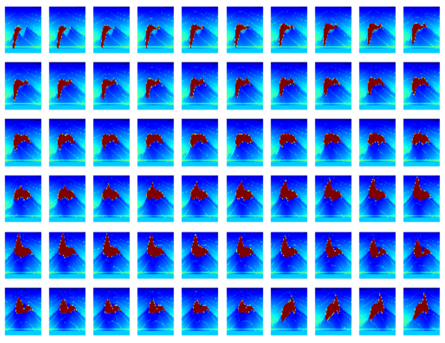}
    \caption{Open data from SEG WIKI.}
    \label{fig:real}
\end{figure}

\begin{figure}
  \centering
    \includegraphics[width=0.8\linewidth]{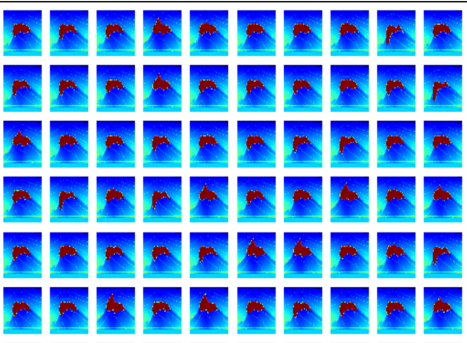}
    \caption{Synthetic data GAN.}
    \label{fig:gan}
\end{figure}

We used the open data from \url{https://wiki.seg.org/wiki/SEG_C3_NA}
to experiment GAN, and the raw data was 3D in nature. Figure \ref{fig:xyz} shows the inline, crossline, and top views. Due to the computational constraints and scope of the project, we limited the whole work in 2D. After around 9000 epochs, we achieved promising results as seen from the comparison between the real data (figure \ref{fig:real}) and synthetic data from GAN (figure \ref{fig:gan}). If in the future, we decide to incorporate the synthetic data to train our model in the following steps, it might be worthy to involve domain experts to comment on the data quality, as nowadays, majority of the quality control for seismic data and images interpretation is still performed by human experts. 

In addition to data generation, we leveraged several metrics to evaluate the performance of machine learning modelling. Structural similarity index measure (SSIM) was adopted to quantitatively evaluate the structural similarity between the targets and predictions. SSIM is a perceptual method based on comparisons on luminanecs, contrasts, and spatial inter-dependency of the local patches. The index value of 1 indicates that two images are identical [25]. Another approach was inspired by [3], which involved capturing the high level features of images based on some pre-trained models, e.g. VGG16. Going beyond the basic metric, we could also redesign the loss functions for general images recovery by combining pixel-to-pixel loss with higher level feature loss.

\pagebreak

\section{Experiments}
\begin{figure}
  \centering
    \includegraphics[width=0.8\linewidth]{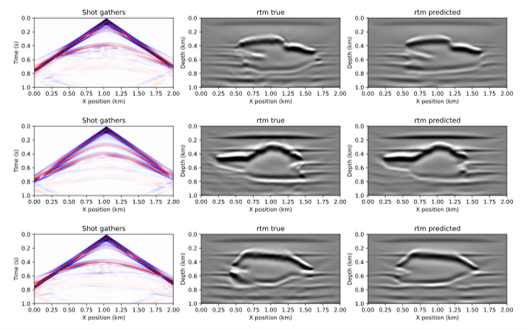}
    \caption{Unet results.}
    \label{fig:Unet}
\end{figure}

\begin{figure}
  \centering
    \includegraphics[width=0.8\linewidth]{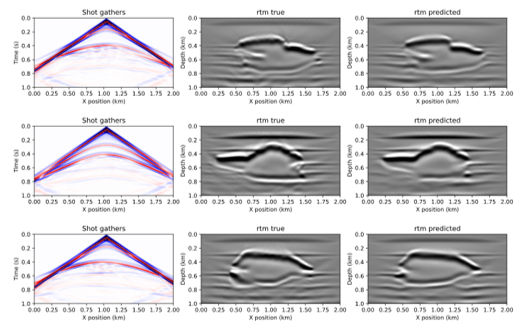}
    \caption{Resnet results.}
    \label{fig:Resnet}
\end{figure}

Once the data was ready (here we use the data generated from the Devito), we kicked off the machine learning experiments by using the wave-fields as input and the RTM images as output to design our end-to-end model. As we all know, the architecture is probably the most important element in building a successful ML model. According to the natures of input and output data, our baseline neural network architecture was built on top of Unet, which simultaneously passes spatial/content information and has been proved to be effective in images segmentation. Traditionally, Unet is formulated to solve segmentation problem, which essentially classifies each pixel within an image. What we did was a little different, as our target domain was subsurface structure, which could be viewed as a regression problem for each pixel in an image. The corresponding loss function had to be re-designed as well. We tested a variety of Unets, and to be more specific, for the encoder part, i.e., the portion from the input to the bottleneck, we replaced it with ResNet34. The motivation of the replacement was that ResNet [3] was one of the leading models in both the “Imagenet” competitions and other use cases in recent years, due to the nature of the residual blocks. With identity shortcut connection, ResNet can solve the gradient vanishing thus network performance degrading issue and provide robust highway network. So we believe the representation capacity of ResNet is able to handle more complex problem, and it could retain better information of the input images. Therefore, we incorporated the ResNet34 structure and used a pre-trained weights (transfer learning) for the encoder part. For the perfromace comparison between Unet and Resnet, please see \ref{fig:Unet}
and \ref{fig:Resnet}. 

\begin{figure}
  \centering
    \includegraphics[width=0.8\linewidth]{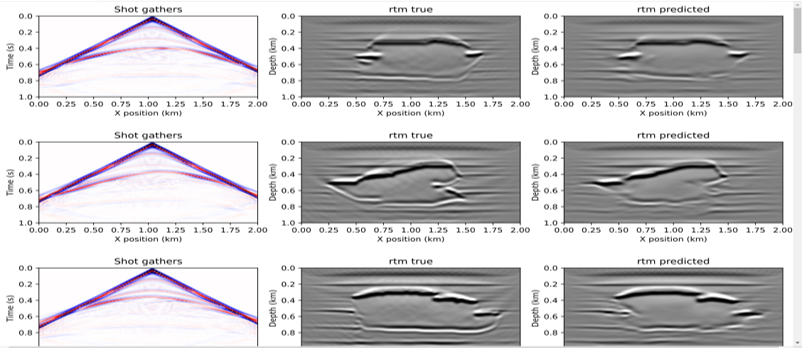}
    \caption{Pixel Loss function.}
    \label{fig:pixel}
\end{figure}

\begin{figure}
  \centering
    \includegraphics[width=0.8\linewidth]{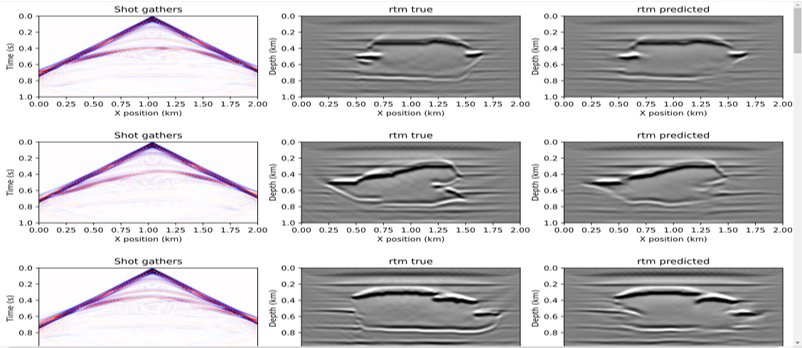}
    \caption{Perception Loss function.}
    \label{fig:perceptual}
\end{figure}

We also elaborated on the loss function dimension, which seems optimistic to pursue further. Pixel-to-pixel loss, e.g., L1 or L2 loss might not be sufficient to capture the structure information of the seismic image, since the loss function should be able to weight pixel differently based on geology information and characteristics. We then borrowed the idea called perceptual loss, from Feifei Li’s paper about style transfer. The following figure \ref{fig:loss} is from the paper [3]. The procedure is to feed the predicted image into the pretrained VGG19 model and extract certain layers’ output and compare them with the corresponding output from the training image into the same VGG19 model. The total loss function is defined as the sum of the L2 norm between each layer output difference. It is based on the idea that shallower layers can extract basic features, and deeper layers can derive high level structure, and combining those together could give a more comprehensive target (regarding images) for the objective function to optimize. We include the how the new loss function (Perceptual loss) improve the training performance, please see \ref{fig:pixel}
and \ref{fig:perceptual}. 

Another approach is through model compression, which could iteratively throw out or zero out unimportant weights in the neural network. Here we experiment with the Level Pruner approach, which is one basic one-shot pruner and one can set a target sparsity level (expressed as a fraction, 0.6 means we will prune 0.6 of the weight parameters). We first sort the weights in the specified layer by their absolute values. And then mask to zero the smallest magnitude weights until the desired sparsity level is reached. One of the impact for model compression is that in the real application, the model is sometimes deployed on the edges, and due to the limitation of equipment and power capacity, deliver a reasonable size of model in the production environment will be critical. 

\begin{figure}
  \centering
    \includegraphics[width=0.8\linewidth]{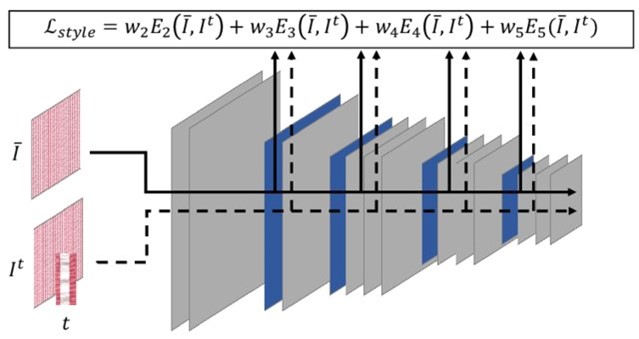}
    \caption{Perception Loss.}
    \label{fig:loss}
\end{figure}

\pagebreak

\section{Architecture Searching and Beyond}
With the baseline Unet and its variations, we have to answer the following questions: how far is the distance from the limit, or how to effectively promote the capacity of the ML model. 

We started with some existing works. Casale et al. discussed the neural architecture search (NAS), where the space of neural network architectures was automatically explored to maximize predictive accuracy for a given task [1]. Despite the success of recent approaches, most existing methods cannot be directly applied to large scale problems because of their prohibitive computational complexity or high memory usage. Then, they proposed a Probabilistic approach to neural ARchitecture SEarCh (PARSEC) that drastically reduced memory requirements while maintaining state-of-the-art computational complexity, making it possible to directly search over more complex architectures and larger data. It’s possible to leverage their work as we explored the best architecture for the seismic inversion problems. We also explored some existing toolkits and platforms to perform the task of searching for the better optimized neural network architecture. One of those toolkits is neural network intelligence (NNI), a lightweight but powerful toolkit to help users in automate feature engineering, neural architecture search, hyperparameter tuning and model compression. We also tested Archai, another NAS platform that allows people to generate efficient deep networks for their applications. Archai aspires to accelerate NAS by enabling easy mix and match between different techniques while ensuring reproducibility, self-documented hyper-parameters, and fair comparison. To achieve that, Archai uses common code base that unifies several algorithms. Archai is extensible and modular to allow a rapid experimentation of new research ideas and develop new NAS algorithms. Archai also hopes to make NAS more accessible to non-experts by providing the powerful configuration system and easy-to-use tools. We might be able to take advantages of those existing tools to speed up our project development. Here we add a few more words about AutoML, which focuses on automating every aspect of the machine learning workflow to increase efficiency and democratize machine learning so that non-experts can apply machine learning to their problems. While AutoML encompasses the automation of a wide range of problems associated with ETL, model training, and model deployment, hyperparameter optimization is the core focus of AutoML. This problem involves configuring the internal settings that govern the behavior of an ML model/algorithm to return a high-quality predictive model. And our problem of NAS can be thought as a smaller problem focusing on finding the optimized backbone or architecture to perform a deep learning task. 

Generally speaking, given a set of search space operation O, we want to find a combination of those operations that maximizes or minimizes the objective function. The idea of searching for a high performing model architecture is not trivial and involves two steps. The first step is searching the cells on a small dataset (e.g. CIFAR10 or CIFAR100), and the second is making model from the searched cells and training it on a big dataset (e.g. ImageNet). A cell can be considered a special block where layers are stacked just like any other models. These cells apply many convolution operations to get feature maps which can be passed over to other cells. A model is made by stacking all the cells in a series to make a complete model. All the papers we mentioned followed a pattern, where two types of cells were searched, namely normal cell and reduction cell.

Among all the possible approaches, we picked differentiable architecture search (DARTS), which is one of the most influential NAS approach. Earlier NAS methods used reinforcement learning or evolution learning which required a large number of computational resources. It could easily take 2000 GPU days of reinforcement learning or 3150 GPU days of evolution. For most organizations, such computing costs are not acceptable. DARTS reduces the search time to 2–3 GPU days, which is quite phenomenal. People could be able to conduct the searching experiments on a single GPU platform such as Colab. DARTS makes the search space continuous by relaxing the categorical choice of a particular operation to a softmax over all possible operations. After relaxation, the goal is to jointly learn the architecture and the weights within all the mixed operations by solving a bilevel optimization problem, which can be posed as to find the mixing probabilities so that validation loss is minimized given weights that are already optimized on the training set. At the end of the search, the architecture is obtained by replacing each mixed operation with the most likely one [10]. 

\begin{figure}
  \centering
    \includegraphics[width=0.8\linewidth]{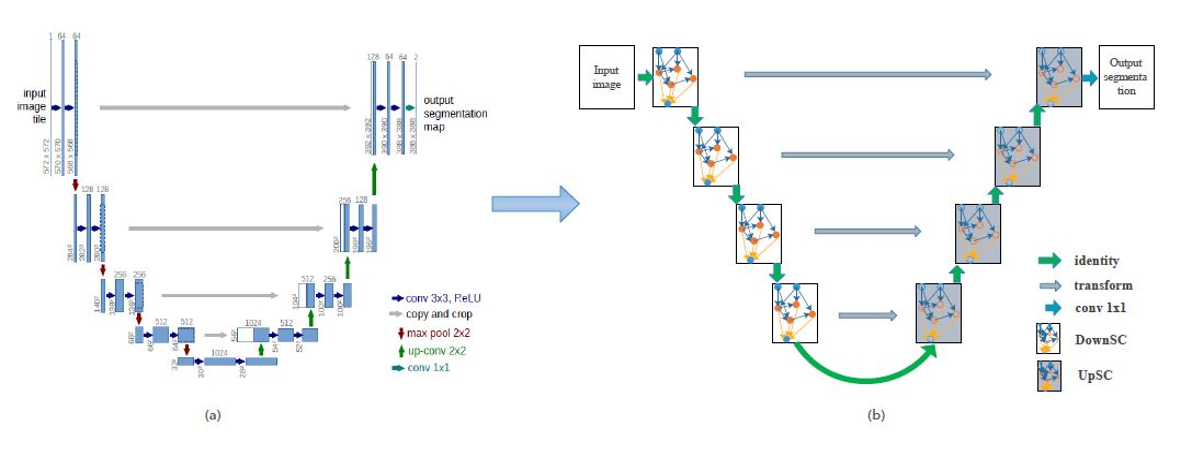}
    \caption{Architectures for Unet (left) and NasUnet (right).}
    \label{fig:newarch}
\end{figure}

\begin{figure}
  \centering
    \includegraphics[width=0.8\linewidth]{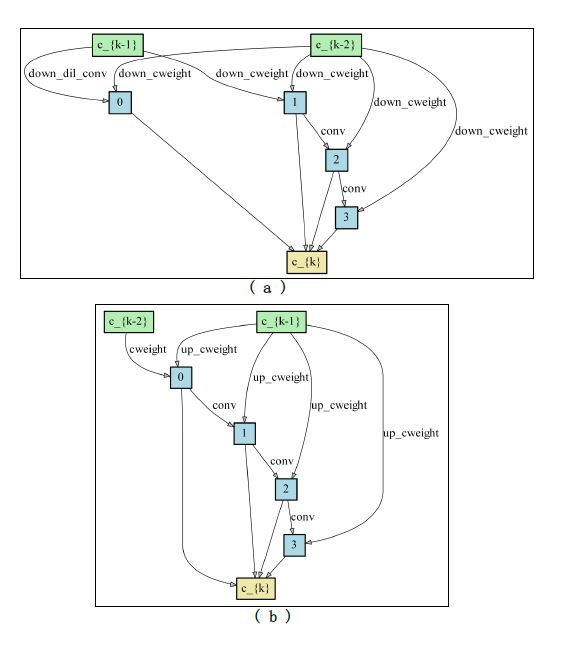}
    \caption{Cell structures from DARTS: (a) encoding cell and (b) decoding cell.}
    \label{fig:cell}
\end{figure}

\begin{figure}
  \centering
    \includegraphics[width=0.8\linewidth]{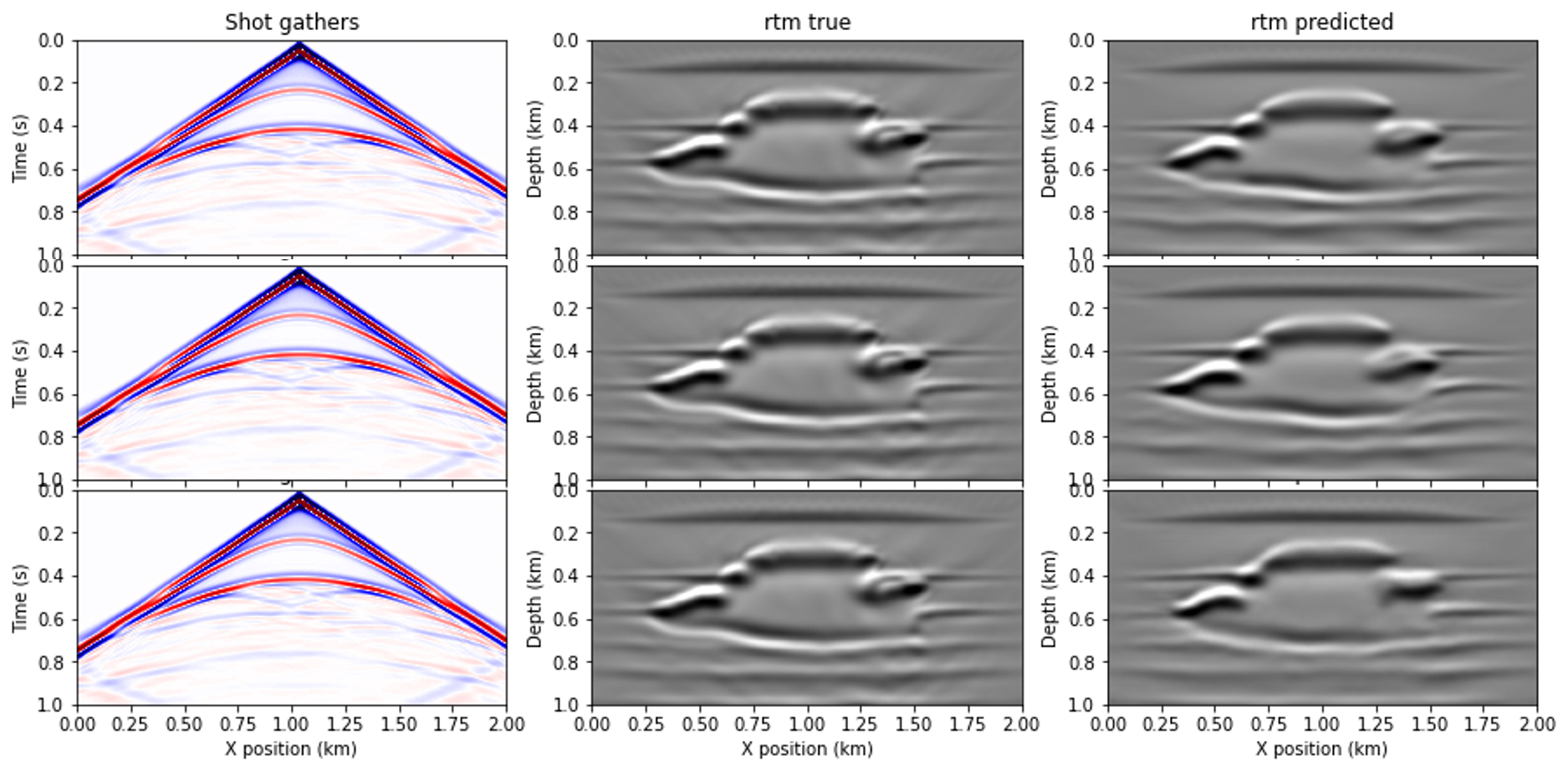}
    \caption{Model comparisons: input wave-field (left), target (middle), and predictions (right) from Unet (top), ResUnet (middle) and NasUnet (bottom).}
    \label{fig:comparisons}
\end{figure}

Figure \ref{fig:comparisons} shows the results from our DARTS architecture, comparing with Unet and Resnet mentioned above. Our search space consisted of 3 dimensions, depth, number of nodes, and operational candidates (Convolution, dilation convolution, max pooling, etc). We used Unet as the backbone, and within each block, searched the optimized structure (see figure \ref{fig:newarch} for details). With more computing power, the backbone structure could also be further optimized as well. Since Unet contains both encoder and decoder, it's therefore natural to split a searching task into the searches of encoding cells and decoding cells. To simplify the problem, we assumed all the encoding cells were same, and the same applied to all the decoding cells. The final NAS cell structures based on Unet (NasUnet) are shown in figure \ref{fig:cell}, a shows the encoder cell, b the decoder cell. The comparison among the predictions using different models (figure \ref{fig:comparisons}) shows that the architecture of NasUnet achieved a competitive performance with Unet and ResUnet. Table \ref{table:1} summarizes the parameter size and SSIM index for three architectures. The SSIM index confirms that, for three models, the structural similarities between the targets and predictions are in the close high level, while the parameter size of NasUnet is much smaller than the other two models. Once the architecture is decided, the training should be much faster for NasUnet.

\pagebreak


\begin{table}[h!]
\centering
\begin{tabular}{||c c c c||} 
 \hline
    & Unet & ResUnet & NasUnet \\ [0.5ex] 
 \hline\hline
Parameter size & $17\times10^6$ & $33\times10^6$ & $0.7\times10^6$\\ 
SSIM & $0.83$ & $0.81$ & $0.77$\\  [1ex] 
 \hline
\end{tabular}
\caption{Comparison in parameter size and SSIM index for different models}
\label{table:1}
\end{table}

\section{Discussion}

In this article, we presented two approaches to generate seismic data, one is physics driven and uses the finite difference method to solve the wave equation and generate the wave-field, and another is GAN that uses the generative network to generate candidates while uses the discriminative network to evaluate the generated candidates. We also formally introduced the metrics, tested different architectures, and evaluated which architecture would be suitable for seismic inversion problems. 

We then tested the ML model and evaluated its performance and limitation based on the data we generated at the first step. The Unet and its variation, ResUnet, were both incorporated in the experiments. Our tests also showed that, besides the model complexity, incorporating higher level features into the cost function also drove the result towards a better state. 

In the architecture searching and beyond effort, we leveraged DARTS to search the better optimized architecture for the inversion problem. The performance of the searched architecture, NasUnet, was comparable with those of Unet and ResUnet, while the model size was much more compressed.


\section*{References}
\small

[1]	Casale, F.P., Gordon, J. and Fusi, N., 2019. Probabilistic neural architecture search. arXiv preprint arXiv:1902.05116. 

[2]	Dong, X. and Yang, Y., 2020. Nas-bench-201: Extending the scope of reproducible neural architecture search. arXiv preprint arXiv:2001.00326.

[3]	Elsken, T., Metzen, J.H. and Hutter, F., 2019. Neural architecture search: A survey. J. Mach. Learn. Res., 20(55), pp.1-21.

[4]	He, K., Zhang, X., Ren, S. and Sun, J., 2016. Deep residual learning for image recognition. In Proceedings of the IEEE conference on computer vision and pattern recognition (pp. 770-778).

[5] Hoffman, J., Tzeng, E., Park, T., Zhu, J.Y., Isola, P., Saenko, K., Efros, A. and Darrell, T., 2018, July. Cycada: Cycle-consistent adversarial domain adaptation. In International conference on machine learning (pp. 1989-1998). PMLR.

[6]	Hu, H., Langford, J., Caruana, R., Mukherjee, S., Horvitz, E.J. and Dey, D., 2019. Efficient forward architecture search. In Advances in Neural Information Processing Systems (pp. 10122-10131).

[7]	Huang, G., Liu, Z., Van Der Maaten, L. and Weinberger, K.Q., 2017. Densely connected convolutional networks. In Proceedings of the IEEE conference on computer vision and pattern recognition (pp. 4700-4708).

[8]	Johnson, J., Alahi, A. and Fei-Fei, L., 2016, October. Perceptual losses for real-time style transfer and super-resolution. In European conference on computer vision (pp. 694-711). Springer, Cham.

[9]	Lange, M., Kukreja, N., Louboutin, M., Luporini, F., Vieira, F., Pandolfo, V., Velesko, P., Kazakas, P. and Gorman, G., 2016, November. Devito: Towards a generic finite difference dsl using symbolic python. In 2016 6th Workshop on Python for High-Performance and Scientific Computing (PyHPC) (pp. 67-75). IEEE.

[10] Liu, H., Simonyan, K. and Yang, Y., 2018. Darts: Differentiable architecture search. arXiv preprint arXiv:1806.09055.

[11] Morlet, J., Arens, G., Fourgeau, E. and Glard, D., 1982. Wave propagation and sampling theory—Part I: Complex signal and scattering in multilayered media. Geophysics, 47(2), pp.203-221.

[12] Nayman, N., Noy, A., Ridnik, T., Friedman, I., Jin, R. and Zelnik, L., 2019. Xnas: Neural architecture search with expert advice. In Advances in Neural Information Processing Systems (pp. 1977-1987).

[13] Pham, H., Guan, M.Y., Zoph, B., Le, Q.V. and Dean, J., 2018. Efficient neural architecture search via parameter sharing. arXiv preprint arXiv:1802.03268.

[14] Ronneberger, O., Fischer, P. and Brox, T., 2015, October. U-net: Convolutional networks for biomedical image segmentation. In International Conference on Medical image computing and computer-assisted intervention (pp. 234-241). Springer, Cham.

[15] Karimpouli, S. and Tahmasebi, P., 2020. Physics informed machine learning: Seismic wave equation. Geoscience Frontiers, 11(6), pp.1993-2001.

[16] Siems, J., Zimmer, L., Zela, A., Lukasik, J., Keuper, M. and Hutter, F., 2020. NAS-Bench-301 and the case for surrogate benchmarks for neural architecture search. arXiv preprint arXiv:2008.09777.

[17] Song, S., Mukerji, T. and Hou, J., 2020. Bridging the gap between geophysics and geology with Generative Adversarial Networks (GANs).

[18] Tan, M. and Le, Q.V., 2019. Efficientnet: Rethinking model scaling for convolutional neural networks. arXiv preprint arXiv:1905.11946.

[19] Weng, Y., Zhou, T., Li, Y. and Qiu, X., 2019. Nas-unet: Neural architecture search for medical image segmentation. IEEE Access, 7, pp.44247-44257.

[20] Yan, Z., Dai, X., Zhang, P., Tian, Y., Wu, B. and Feiszli, M., 2020. FP-NAS: Fast Probabilistic Neural Architecture Search. arXiv e-prints, pp.arXiv-2011.

[21] Yang, F. and Ma, J., 2019. Deep-learning inversion: A next-generation seismic velocity model building method. Geophysics, 84(4), pp.R583-R599.

[22] Ying, C., Klein, A., Christiansen, E., Real, E., Murphy, K. and Hutter, F., 2019, May. Nas-bench-101: Towards reproducible neural architecture search. In International Conference on Machine Learning (pp. 7105-7114). PMLR.

[23] Ying, C., Klein, A., Christiansen, E., Real, E., Murphy, K. and Hutter, F., 2019, May. Nas-bench-101: Towards reproducible neural architecture search. In International Conference on Machine Learning (pp. 7105-7114). PMLR.

[24] Zoph, B. and Le, Q.V., 2016. Neural architecture search with reinforcement learning. arXiv preprint arXiv:1611.01578.

[25] Wang, Z., Conrad, A., Sheikh, H., and Simoncellim E., 2004, April. Image Quality Assessment: From Error Visibility to Structural Similarity. IEEE Transactions on Image Processing, 13(4), pp. 600 - 612.

\end{document}